\pdfoutput=1
\documentclass[11pt]{article}

\usepackage[]{ACL2023}

\usepackage{times}
\usepackage{latexsym}

\usepackage[T1]{fontenc}
\usepackage{graphicx}
\usepackage{multirow}
\usepackage{hyperref}
\usepackage[utf8]{inputenc}
\usepackage {amsfonts,amssymb} 
\usepackage{amsmath}

\usepackage{microtype}

\usepackage{inconsolata}

\title{CFSum: A Coarse-to-Fine Contribution Network for \\
Multimodal Summarization}

\author{Min Xiao\textsuperscript{ 1,2}, Junnan Zhu\textsuperscript{1,2}, Haitao Lin\textsuperscript{1,2}, Yu Zhou\textsuperscript{1,3}\thanks{\ \ Corresponding author.},  Chengqing Zong\textsuperscript{1,2}\\
\textsuperscript{1} State Key Laboratory of Multimodal Artificial Intelligence Systems, \\
Institute of Automation, CAS, Beijing, China \\
\textsuperscript{2} School of Artificial Intelligence, University of Chinese Academy of Sciences, Beijing, China \\
\textsuperscript{3} Fanyu AI Laboratory, Zhongke Fanyu Technology Co., Ltd, Beijing, China\\
$\left\{\right.$min.xiao, junnan.zhu, haitao.lin, yzhou,  \\
cqzong$\left .\right\}$@nlpr.ia.ac.cn,
}
\begin{document}
\maketitle
\begin{abstract}
Multimodal summarization usually suffers from the problem that the contribution of the visual modality is unclear. Existing multimodal summarization approaches focus on designing the fusion methods of different modalities, while ignoring the adaptive conditions under which visual modalities are useful. Therefore, we propose a novel \textbf{C}oarse-to-\textbf{F}ine contribution network for multimodal \textbf{Sum}marization (CFSum) to consider different contributions of images for summarization. First, to eliminate the interference of useless images, we propose a pre-filter module to abandon useless images. Second, to make accurate use of useful images, we propose two levels of visual complement modules, word level and phrase level. Specifically, image contributions are calculated and are adopted to guide the attention of both textual and visual modalities. Experimental results have shown that CFSum significantly outperforms multiple strong baselines on the standard benchmark. Furthermore, the analysis verifies that useful images can even help generate non-visual words which are implicitly represented in the image\footnote{Code is available at \url{https://github.com/xiaomin418/CFSum}}.
\end{abstract}

\section{Introduction}
With the information explosion, the internet is flooded with various multimodal information. Multimodal summarization (MMS) can help generate more abundant and comprehensive summary information than unimodal based on extra visual information. Existing studies on multimodal summarization include multimodal sentence summarization \citep{ijcai2018-577}, multimodal summarization with multimodal output \citep{zhu-etal-2018-msmo}, multimodal meeting summarization \citep{li-etal-2019-keep} and so on. In this paper, we focus on the task that generating a text summary based on the input of a text and an image. It has been proved that integrating multimodal data can help improve the quality of the summary \citep{ijcai2018-577, 10.1007/978-3-030-45442-5_24, palaskar-etal-2019-multimodal, yu-etal-2021-vision}.
\begin{figure}[t]
    \centering
    \includegraphics[width=\linewidth,scale=1.00]{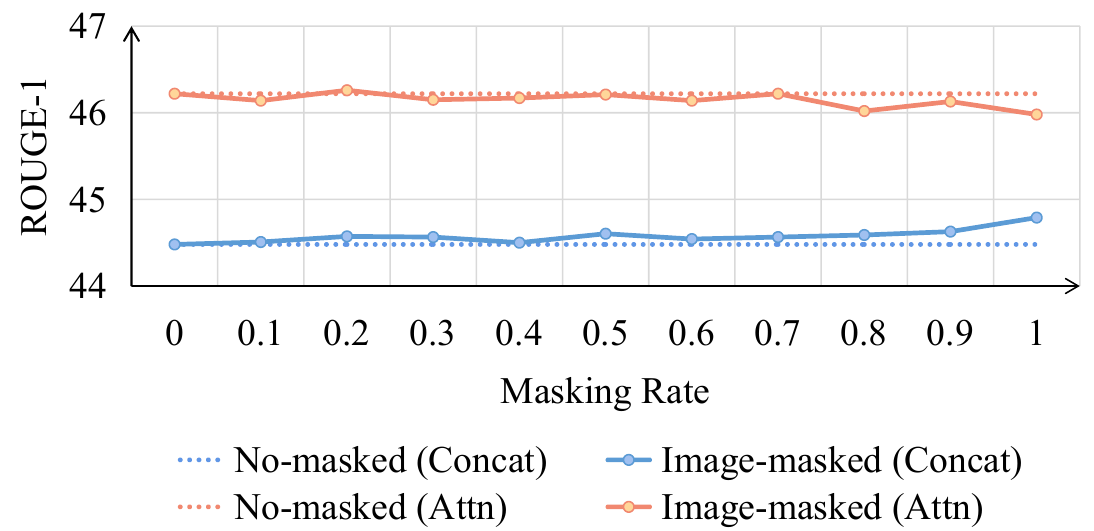}
    \caption{Experiments on existing mainstream multimodal summarization models. The performance is not affected by masking images. ``Concat'' is the concatenate fusion method, and ``Attn'' is the attention-based fusion method.}
    \label{fig:introduction}
\end{figure}
\par However, it is unclear whether the visual modality can indeed benefit the process of summarization. Thus, we conduct an experiment to explore the influence of masking images on the summary. As shown in Figure \ref{fig:introduction}, the solid lines mean the performance of summary generated by masking portions of images, and the dashed lines indicate the origin performance. It can be observed that the dashed and the solid lines roughly coincide, which indicates that masking images do not affect the performance of the multimodal model. Some masking rates can even raise the ROUGE-1 value of the summary. It indicates that existing models do not make effective use of image information for the summary.
\par Existing approaches have two major problems. First, existing studies focus on multimodal fusion, such as concatenate, attention-based, and gate-based fusion (referring to \nameref{sec:related}). However, they ignore the adaptive conditions under which visual modalities are helpful. Thus they are poor at extracting useful visual information. Furthermore, all fusion methods do not explicitly model the image complementarity for the summary. Especially for the attention-based method, the inter-attention is not accurate enough, which leads to inefficient use of the image. Second, in many samples, the image may introduce noise, while existing fusion methods assume that all images are helpful for the summary without considering the interference of useless images. As analyzed above, we believe that: 1) It is essential to eliminate the influence of the useless image. 2) The contributions of the image to the summary need to be clarified. In particular, it is necessary to consider the complementarity of visual information relative to textual information. 
\par Although we notice the lack of image contributions, it is difficult to detach various roles of images from a single fusion layer. Thus, in this work, we propose a novel Coarse-to-Fine contribution network for multimodal Summarization (CFSum) to extract the role of the image at different stages. First, we apply a pre-filter module to abandon useless images. It coarsely obtains helpful images for the summary. Specifically, the consistency of content between image and text is calculated. If the consistency is low, the image will be masked in subsequent encoding. Second, when the image is coarsely useful, the complement module is employed to finely guide the fusion of text with the image. To consider image contributions for text with different granularities, the complement module consists of two levels, word level and phrase level. For the word level complement module, to obtain the image complementarity over the text, the difference between bi-modal and uni-modal inputs is measured through a classification task. Then we add a loss to guide the attention between words and the image. For the phrase level complement module, similar to the word level, the image complementarity on phrases is acquired to guide the attention between phrases and the image. Through these modules, the model can acquire more explicit image contributions and provide better multimodal encoding for summary generation.
\par Our contributions are as follows: 
\par (1) We propose a Coarse-to-Fine contribution network for multimodal Summarization (CFSum) to model different contributions of images for summarization.
\par (2) We innovatively design a pre-filter module to coarsely reduce the interference of the useless images and develop two visual complement modules to finely obtain image complementarity over the summary.
\par (3) Experimental results show that our model outperforms strong baselines. Besides, extensive analysis proves that useful image even contributes to non-visual words which are implicitly represented in the image.

\section{Related Work}
\label{sec:related}
\paragraph{Multimodal Summarization Tasks.}
In the field of multimodal summarization, there are usually three steps. First, different feature extractor modules are adopted to extract the features of the text and the image, respectively. Second, the different features are fused at the fusion layer. Finally, the fused context features are fed into the text decoder to generate a summary. 
Existing studies focus on multimodal fusion. Specifically, the fusion methods consist of concatenate, attention-based, and gate-based. The concatenate fusion directly concatenates multimodal features into a fusion context \citep{ijcai2018-577, Li_Yuan_Xu_Wu_He_Zhou_2020}. It can fully extract high-level features of different modalities, but there is a large gap between high-dimensional spaces. The attention-based methods fuse all multimodal features with attention mechanism \citep{10.1016/j.knosys.2021.107152, palaskar-etal-2019-multimodal, 9947020}, which can get the correlations between each unit of text and image. Gate-based methods take text as the central modality \citep{DBLP:journals/corr/abs-2109-05199} and exploit images to help focus on the core information \citep{liu-etal-2020-multistage, li-etal-2020-multimodal}. In summary, (1) all fusion methods do not explicitly model the image complementarity for the summary, which leads to inefficient use of the image. (2) concatenate and attention-based cannot eliminate the influence of useless images in the fusion layer.
\begin{figure*}[t]
    \centering
    \includegraphics[width=0.9\linewidth,scale=1.00]{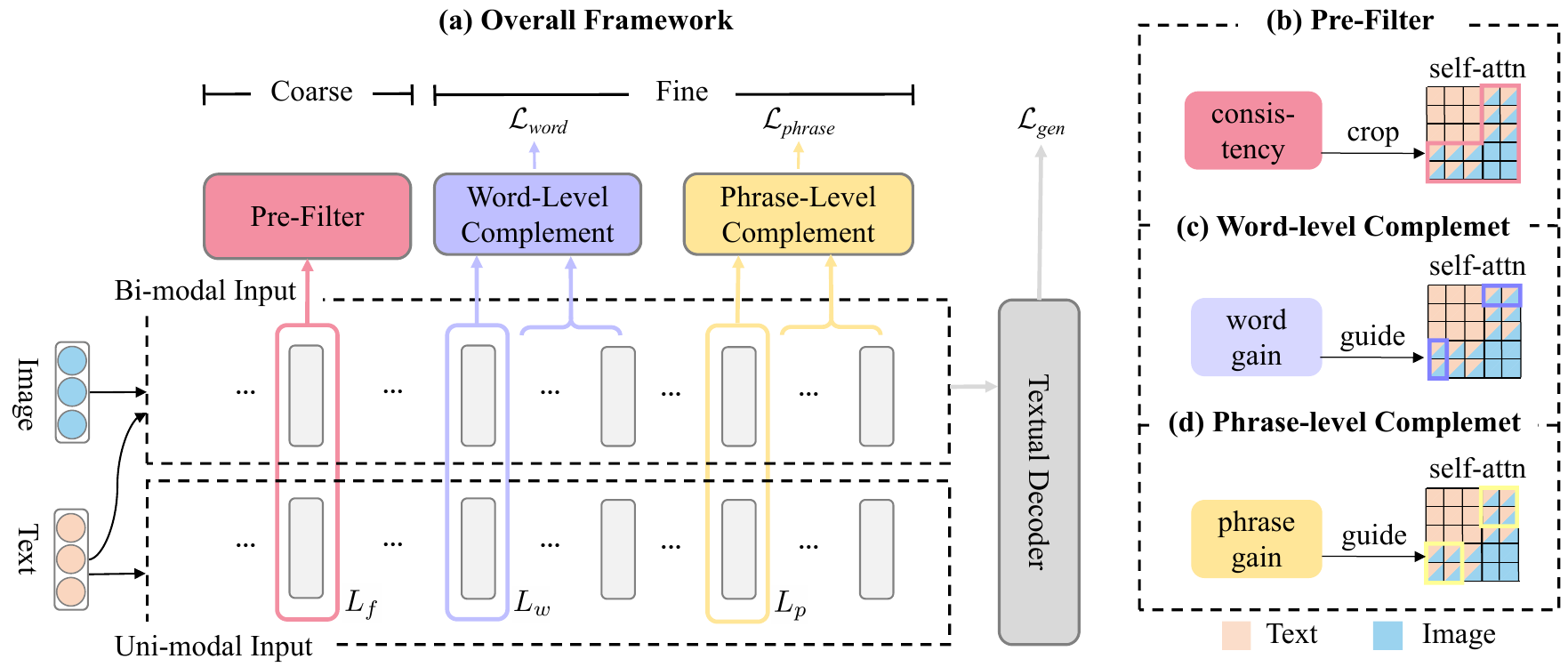}
    \caption{CFSum framework. $L_f, L_w, L_p$ denote the starting layer of the pre-filter, the word-level complement, and the phrase-level complement modules, respectively.}
    \label{fig:HCG-MMS}
\end{figure*}
\paragraph{Cross-modal tasks.}  
Some studies have noted the contributions of modalities and explored the cross-modal influence in 
 other multimodal tasks. \citet{zeng-etal-2021-making-contribution} propose loss modulation to explore the contribution of individual modalities and devise a modality filter to reduce modality noise, which considers consistency and complementarity between different modalities. \citet{zhu-etal-2018-msmo} propose multi-task summarization: the method also selects the image that best matches the summary when generating a text summary. It guarantees the positive effect of images on the summary. \citet{e24060764} exploit ReLu-based cross-attention to align visual features to textual representation, which abandons low-value attention scores for those unaligned visual features. Inspired by the above studies, we propose CFSum, which considers various image contributions for better encoding input text and generating the final summary.

\section{Proposed Methods}
\subsection{Overview}
\label{sec:overview}
In this section, we introduce the details of CFSum. Given a dataset consisting of $n$ triplets $(t_i, v_i,s_i)_{i \in [1,n]}$ with a text $t_i$, an image $v_i$, and a summary $s_i$, the multimodal summarization task aims at generating $s_i$ based on $t_i$ and $v_i$.
\par As depicted in Figure \ref{fig:HCG-MMS}, the CFSum takes bi-modal and uni-modal streams as input parallelly. It builds coarse and fine image contributions with three modules (\nameref{sec:hierachical}). First, the pre-filter module coarsely filters the images inconsistent with texts (\nameref{sec:filtering}). Second, two levels of visual complement modules consisting of word level (\nameref{sec:word-level}) and phrase level (\nameref{sec:semantic-level}) make accurate use of useful images.

\subsection{Coarse-to-Fine Structure}
\label{sec:hierachical}
We build our model based on the multimodal transformer UNITER \citep{10.1007/978-3-030-58577-8_7} and GRU \citep{https://doi.org/10.48550/arxiv.1412.3555} encoder-decoder architectures. We refer the model to UniG. As shown in Figure \ref{fig:HCG-MMS}(a), in order to evaluate the complementarity of different modalities, the bi-modal and uni-modal inputs are operated parallelly with the same encoder. The two parallel streams can catch the gain of the image. Additionally, we generate a summary relying on bi-modal encoding. Uni-modal encoding assists in measuring various contributions and guiding the bi-modal encoding. 
\par Specifically, the multimodal encoder consists of $L=12$ multimodal transformer layers. We serve the $L$ layers as a hierarchical structure and divide $L$ layers into three parts as shown in Figure \ref{fig:HCG-MMS}(a). $L_f, L_w, L_p$ mark as the starting layer of the pre-filter, the word-level complement, and the phrase-level complement modules, respectively. Existing studies assume all images benefit summary generation or input text encoding, resulting in damage from unnecessary images. The pre-filter module is utilized to eliminate the interference of misleading images in advance. Next, the word-level complement module is developed to model the gain of the image on input words for the summary. Then the image gain guides the subsequent attention between words and the image. Finally, similar to the word level, the phrase-level complement module concentrates on phrases at higher layers. Each component will be elaborated in the following sub-sections.
 
\subsection{Pre-filter Module}
\label{sec:filtering}
The bi-modal and uni-modal features from the $i^{th}$ layer are encoded as $m^i\in \mathbb{R}^{C \times H}, {u}^i\in \mathbb{R}^{T \times H}$, where $i\in [1,L]$, and $C,T$ denote the lengths of bi-modal and uni-modal tokens. $H$ denotes the hidden dimension. The bi-modal self-attention matrix in the $i^{th}$ layer is $A^i=(a_{r,s}^i) \in \mathbb{R}^{C \times C}$.

\par The pre-filter module aims at filtering images that are unnecessary to the summary. As shown in Figure \ref{fig:HCG-MMS}(a), given two encoded features ${m}^{L_f}$ and ${u}^{L_f}$ from the $L_f^{th}$ layer, the goal of the filtering module is to select those useless images and guide the self-attention of all subsequent layers. We believe that if the bi-modal feature has low consistency with the uni-modal feature, the image may introduce interferential information. Specifically, we first calculate the consistency $\Delta^{C}$ between uni-modal feature $u^{L_f}$ and bi-modal feature $m^{L_f}$ as follows:

\begin{small}
\begin{align}
&pu = \textrm{MeanPool}(u^{L_f}), \\
&pm = \textrm{MeanPool}(m^{L_f}), \\
&\Delta^{C} = \textrm{Sign}(\textrm{cosine}(pu, pm)-\alpha)
\end{align}
\end{small}%
We define the indicator function as:
\begin{small}
\begin{equation}
    I_{r,s}=
   \begin{cases}
   1,&\mbox{$r\geq T,s\geq T$}\\
   0,&\mbox{otherwise}
   \end{cases}
  \end{equation}
\end{small}%
which represents the text attending to the image, the image attending to the text, and the image attending to itself shown in Figure \ref{fig:HCG-MMS}(a). Then we calculate the new subsequent self-attention $na^i_{r,s}$ with:

\begin{small}
    \begin{align}
        na^i_{r,s}=&a^i_{r,s}\times(1-I_{r,s}) + a^i_{r,s}\times I_{r,s}\times\Delta^{C},\\
        &i\in [L_f+1,L]\nonumber
    \end{align}
\end{small}%
By correcting the attention matrix, the image with a large deviation in content is cropped out. In other words, the multimodal inconsistency features degenerate into text-only features through this process. The simple method has been shown to be effective in our experiments.

\subsection{Word-level Complement}
\label{sec:word-level}
This section introduces a word-level complement module, considered as an auxiliary task during the training process. First, we measure the image gain on input words for the summary. Then the image gain is applied to guide the attention between words and the image (as shown in Figure \ref{fig:HCG-MMS}(b)).
\paragraph{Image gain measurement.} Intuitively, the text tokens should concern the image which is helpful for the summary. In previous attention-based studies, inter-modality correlation can be modeled as $\textrm{softmax}(\frac{QK}{\sqrt{D}})V$. $Q, K, V$ are the projected features from the bi-modal input. However, it does not explicitly model the image complementarity for the summary, which leads to inefficient use of the image.
\par Following the motivation above, we hope to calculate the image gain on the summary with mutual information. In other words, we want to measure whether generating summaries based on bi-modal feature $m^L$ is more deterministic than generating summaries based on uni-modal feature $u^L$. Thus, we expect to calculate the image gain on the $k$-th word of the reference summary:

\begin{small}
    \begin{equation}
        GI_k = Gain(s_k/u^L, s_k/m^L)
    \end{equation}
\end{small}%
\par However, we intend to obtain $GI_k$ before generating summary $S$ and encoding $m^L$. Thus $GI$ can be beneficial for generating $S$ and encoding $m^L$. To this end, we define Copy Classification task $Y$ to approximate the summary task $S$: for each input text token $t_j$, the target is to binary categorize whether it appears in the reference summary. If the token appears in the reference summary, it is classified as $\hat{y_j}=1$; otherwise, $\hat{y_j}=0$. Next, the $GI_j$ is given by:

\begin{small}
    \begin{equation}
        GI_j = Gain(y_j/u^{L_w}, y_j/m^{L_w})\label{fomula:gi_y}
    \end{equation}
\end{small}%
where $u^{L_w},m^{L_w}$ denote the uni-modal and the bi-modal feature acquired by $L_w^{th}$ layer. Finally, we measure the gain that the image brings to predict whether a word appears correctly in the summary as follows:

\begin{small}
    \begin{align}
        GI_j & = Gain(y_j/m^{L_w}, y_j/u^{L_w})\nonumber \\
            & = logP(y_j=\hat{y_j}/m^{L_w})-logP(y_j=\hat{y_j}/u^{L_w}) \label{fomula:gi_log}
    \end{align}
\end{small}%
Derivation details can refer to the Appendix \ref{appendix:derivation}. In addition, to ensure the correct gain direction, we add a binary cross-entropy loss to train the Copy Classification Task $Y$: 

\begin{small}
    \begin{equation}
        \mathcal{L}_{copyc} = \textrm{BCE}(y_j,\hat{y_j}/m^{L_w})+\textrm{BCE}(y_j,\hat{y_j}/u^{L_w})
    \end{equation}
\end{small}%

\paragraph{Image gain application.} We introduce divergence loss to restrain that the image with greater gain should receive more textual attention. In successive $i^{th}\,_{i\in [L_w+1,L_w+3]}$ layer, the average inter-attention between each text token $t_j$ and the image is:

\begin{small}
    \begin{equation}
T2V_j^i = \frac{1}{2(C-T)}(\sum_{s=T+1}^{s=C} a_{j,s}^{i}+\sum_{s=T+1}^{s=C}a_{s,j}^{i}) \label{fomula:T2V}
    \end{equation} 
\end{small}% 
where $a_{j,s}^{i}, a_{s,j}^{i}$ represent the attention of image-to-text and text-to-image, respectively. 
\par Finally, an attention divergence loss is added to restrain the inter-attention scores $T2V_j^i$ with $GI_j$:

\begin{small}
\begin{equation}
    \mathcal{L}_{word}=\textrm{KL}(\textrm{Softmax}(GI_j)||\textrm{Avg}(T2V_j^i))
\end{equation}
\end{small}%
By minimizing the divergence loss, the text token attends to the image according to the gain it brings. Interaction between word gain and inter-attention learns to pay attention to the useful image. Appendix \ref{appendix:exs} provides examples to figure out the word-level complement.

\subsection{Phrase-level Complement}
\label{sec:semantic-level}
Considering the image contribution to text of different granularities, we put forward a phrase-level complement module similar to the word level (as shown in Figure \ref{fig:HCG-MMS}(c)).
\paragraph{Image gain measurement.}  Different from copy classification task at the word level, we define Copy Scorer task to measure the image gain on phrases: We obtain phrases $\{p_1,...,p_k...\}$ from the text with StandfordNLP\footnote{\url{ https://github.com/stanfordnlp}}. $\{l_1,...,l_k...\}$ is the number of words in the phrases. The task targets scoring the proportion of words that appear in both the phrase and the reference summary:

\begin{small}
    \begin{align}
        & R_{p_k}^u=\textrm{Scorer}(u^{L_p})\\
        & R_{p_k}^m=\textrm{Scorer}(m^{L_p})
    \end{align}
\end{small}%
where Scorer is a MLP. The ground truth proportion is obtained with the following:

\begin{small}
    \begin{equation}
        \hat{R_{p_k}}=\frac{Count_{\,t_{j'}\in p_k}(t_{j'})}{l_k}
    \end{equation}
\end{small}%
where $Count_{\,t_{j'}\in p_k}$ denotes the number of words that appear in both the phrase $p_k$ and the reference summary. Therefore, the image gain on phrase can be acquired as:

\begin{small}
    \begin{equation}
        GS_{p_k}=|R_{p_k}^u-\hat{R_{p_k}}| -|R_{p_k}^m-\hat{R_{p_k}}|
    \end{equation}
\end{small}%
Similarly, to guarantee the correctness of phrase gain, we add a squared loss for the Copy Scorer task:

\begin{small}
    \begin{equation}
        \mathcal{L}_{copys} = \textrm{MSE}(R_{p_k}^m,\hat{R_{p_k}})
        +\textrm{MSE}(R_{p_k}^u,\hat{R_{p_k}})
    \end{equation}
\end{small}%
Especially, for the convenience of applying phrase gain $GS_{p_k}$, we project it to token gain $GS_{j}$ as:

\begin{small}
    \begin{equation}
        GS_j=\max \{GS_{p_k},t_j \in p_k\}
    \end{equation}
\end{small}%
\paragraph{Image gain application.} Second, we introduce a phrase attention divergence loss to restrain that the image with greater phrase gain should receive more textual attention. We obtain the inter-attention score $T2V_j^i$ from $i\in [L_p+1,L_p+3]$ layers as formula \ref{fomula:T2V}. Finally, we restrain it with the following:

\begin{small}
\begin{equation}
    \mathcal{L}_{phrase}=\textrm{KL}(\textrm{Softmax}(GS_j)||\textrm{Avg}(T2V_j^i))
\end{equation}
\end{small}%
The phrase-level restraint guarantees the image contributing to the text of phrase granularity.

\subsection{Training and Inference}
\label{sec:objectives}
In the training phase, to ensure the accuracy of the information difference between bi-modal and uni-modal, we initialize the model only with the summary generation loss. We apply negative log-likelihood for the target word sequence as the overall loss:

\begin{small}
    \begin{equation}
       \mathcal{L}_{gen} = \frac{1}{T}\sum_{t=1}^T(-\log{P(s_t)})
    \end{equation}
\end{small}%
Then the model is finetuned with the hierarchical modules' objectives:

\begin{small}
    \begin{equation}
        \mathcal{L} = \mathcal{L}_{gen} + \mathcal{L}_{word} + \mathcal{L}_{phrase}+ \mathcal{L}_{copyc} + \mathcal{L}_{copys}
    \end{equation}
\end{small}%
\par In the inference phase, we only maintain the pre-filter module. $\mathcal{L}_{word}$ and $\mathcal{L}_{phrase}$ are added to let the model learn how to fuse multimodal information. Hence, differences in training and inference phases would not hurt the generation.

\begin{table}[ht]
\centering
\resizebox{0.9\linewidth}{!}{%
\begin{tabular}{l|rcc}
\hline
Dataset 
& \multicolumn{1}{c}{Size} 
& \multicolumn{1}{l}{Src. Length} 
& \multicolumn{1}{l}{Ref. Length} \\ 
\hline
train & 62,000 & 11/21.68/63 & 2/7.72/25 \\
dev & 2,000 & 11/24.35/47 & 3/7.68/17 \\
test & 2,000 & 11/22.97/51 & 3/7.67/24 \\ 
\hline
\end{tabular}%
}
\caption{Statistical information about the dataset. ``Src. Length'' and ``Ref. Length'' denote the number of words in the source sentence and reference summary. Three values in each column represent: Min, Avg, Max.}
\label{tab:dataset}
\end{table}

\section{Experiment}
\subsection{Settings}
We experiment with the multimodal sentence summarization dataset\footnote{\url{http://www.nlpr.ia.ac.cn/cip/dataset.htm}} \citep{Li2018MultimodalSS}. It contains 66,000 samples in total. 
\begin{table*}[t]
\centering
\resizebox{0.8\linewidth}{!}{%
\begin{tabular}{cl|cccccc}
\hline
\multicolumn{2}{c|}{} & \multicolumn{1}{l}{ROUGE-1} & \multicolumn{1}{l}{ROUGE-2} & \multicolumn{1}{l}{ROUGE-L} & \multicolumn{1}{l}{BLEU} & \multicolumn{1}{l}{BERTScore} & \multicolumn{1}{l}{MoverScore} \\ \hline
\multicolumn{2}{l|}{Lead$\vartriangle$} & 33.64 & 13.40 & 31.84 & - & - & - \\
\multicolumn{2}{l|}{Compress$\vartriangle$} & 31.56 & 11.02 & 28.87 & - & - & - \\ 
\multicolumn{2}{l|}{ABS$\vartriangle$} & 35.95 & 18.21 & 31.89 & - & - & - \\ 
\multicolumn{2}{l|}{SEASS$\vartriangle$} & 44.86 & 23.03 & 41.92 & - & - & - \\
\multicolumn{2}{l|}{Multi-Source$\vartriangle$} & 39.67 & 19.11 & 38.03 & - & - & - \\
\multicolumn{2}{l|}{Doubly-Attention$\vartriangle$} & 41.11 & 21.75 & 39.92 & - & - & - \\
\multicolumn{2}{l|}{MAtt$\vartriangle$} & 47.28 & 24.85 & 44.48 & - & - & - \\
\multicolumn{2}{l|}{MSE$\vartriangle$} & 45.63 &  23.68 & 42.97 & - & - & - \\ 
\hline
\multicolumn{2}{l|}{UniG (T)} & 45.90 & 24.08 & 42.98 & 47.09 & 86.54 & 31.06 \\
\multicolumn{2}{l|}{UniG} & 46.22 & 24.28 & 43.47 & 46.85 & 86.57 & 30.95 \\ \hline
\multirow{3}{*}{K1} & CFSum-F$_3$ & 47.39* & 25.42* & 44.35* & 48.51* & 86.90* & 31.89* \\
 & CFSum-W$_6$ & 47.33* & 25.38* & 44.26* & 48.43* & 86.91* & 31.84* \\
 & CFSum-P$_9$ & 47.28* & 25.13* & 44.18* & 48.19* & 86.91* & 31.67 \\ \hline
\multirow{3}{*}{K2} & CFSum-W$_6$F$_9$ & 47.53* & 25.37* & 44.41* & 48.48* & 86.94* & 32.24* \\
 & CFSum-F$_3$W$_6$ & 47.66* & 25.33* & 44.54* & 48.45* & 86.95* & 31.88* \\
 & CFSum-F$_3$P$_9$ & 47.72* & 25.51* & 44.58* & 48.66* & 86.96* & 32.03* \\ \hline
\multirow{2}{*}{K3} & CFSum-F$_3$W$_6$P$_9$ & \textbf{47.86}* & \textbf{25.64}* & \textbf{44.64}* & \textbf{48.83}* & \textbf{86.98}* & \textbf{32.36}* \\
 & CFSum-F$_9$W$_3$P$_6$ & 47.58* & 25.42* & 44.49* & 48.35* & 86.95* & 32.10* \\ \hline
\end{tabular}%
}
\caption{Automatic evaluation results of CFSum. ``$\vartriangle$'' marks the results from \citet{ijcai2018-577} and \citet{li-etal-2020-multimodal}\protect\footnotemark. ``K1/2/3'' denotes one/two/three kind(s) of contribution(s). ``*'' indicates the model performs significantly better than the UniG by the 95\% confidence interval (p<0.05).}
\label{tab:experiment}
\end{table*}%
And each sample is a triplet of <sentence, image, summary>. Some statistical information is shown in Table \ref{tab:dataset}. Appendix \ref{appendix:dataset} gives the categories of test images.

% \usepackage{graphicx}

\footnotetext{Because there is no output from these systems, we only report ROUGEs in papers. In addition, BLEU, BERTScore, and MoverScore cannot be recalculated.}
\par We set both the text embedding dimension and hidden dimension as 768. We apply ``bert-base-uncased'' \citep{devlin-etal-2019-bert} vocabulary with 28,996 tokens. The dropout \citep{10.5555/2627435.2670313} rate is set to $0.1$. Besides, the batch size is set to $8$. For texts, we use the max text encoding length of 60, and the minimum text decoding length is 8. For images, the object detection tool BUTD \citep{8578734} is applied to extract the image feature, with the maximum boxes as $36$. We use the Adam \citep{kingma2014adam} optimizer and set the learning rate as $5e-05$, momentum parameters as $\beta_1=0.9, \beta_2=0.98$. 
The model is initially trained with the summary generation loss for 35 epochs. To obtain our final model, we train for a further 15 epochs with the hierarchical framework. In the test phase, we employ beam search and set the beam size as $4$ to generate the summary. The parameter $\alpha$ in the pre-filter module is set as $\alpha=0.65$.

\subsection{Comparative Methods}
\textbf{Lead:} Exploiting the first eight words as the summary.\\
\textbf{Compress} \citep{Clarke08globalinference}:  It uses integer linear programming to infer global optimal compressions. \\
\textbf{ABS} \citep{rush-etal-2015-neural}: It utilizes an attention-based model to generate words of summary conditioned on the input text.\\
\textbf{SEASS} \citep{zhou-etal-2017-selective}: It constructs a second-level sentence representation with a sentence encoder and a selective gate for summarization. \\
\textbf{Multi-Source} \citep{libovicky-helcl-2017-attention}: It combines multiple source modalities based on the hierarchical attention mechanisms over each modality for solving the multimodal machine translation.\\
\textbf{Doubley-attentive} \citep{calixto-etal-2017-doubly}: It uses two separate attention mechanisms to incorporate the visual feature, which minified the gap between the image and the translation.\\
\textbf{MAtt} \citep{ijcai2018-577}: It proposes modality attention and image filtering for multimodal summarization.\\
\textbf{MSE} \citep{li-etal-2020-multimodal}: It proposes to apply the visual selective gates to multimodal summarization.\\
\textbf{UniG}: It is our base model with multimodal transformer UNITER and GRU decoder.\\
\textbf{UniG (T)}: UniG fed only with textual modality.

\begin{figure*}[t]
    \centering
    \includegraphics[width=0.9\linewidth,scale=1.00]{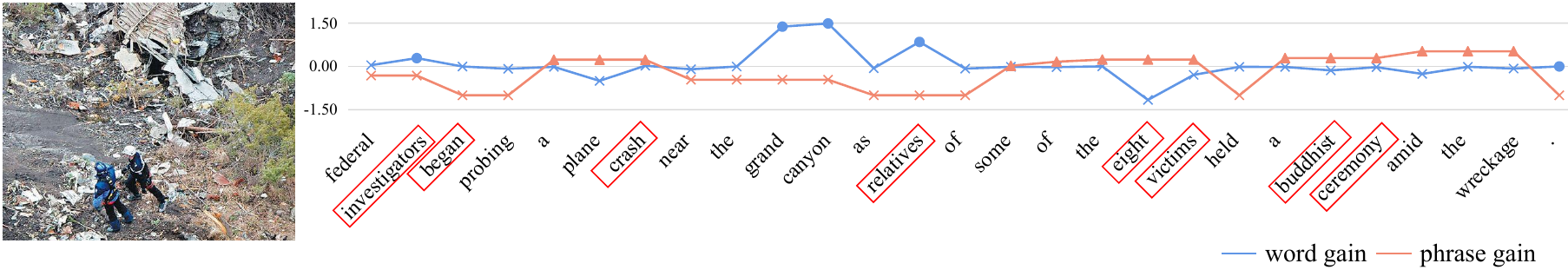}
    \caption{Visualization of word-complement gain and phrase-complement gain produced by our model. \textcolor[RGB]{240,136,113}{$\blacktriangle$}/\textcolor[RGB]{96,150,230}{$\bullet$} indicates that the value is greater than 0.}
    \label{fig:analysis}
\end{figure*}

\subsection{Automatic Evaluation Results}
Our methods are reported with six automatic metrics, including ROUGE-1, ROUGE-2, ROUGE-L \citep{lin-hovy-2002-manual}, BLEU \citep{papineni-etal-2002-bleu}, BERTScore \citep{bert-score}, and MoverScore \citep{zhao-etal-2019-moverscore}. More details of evaluation scripts are given in Appendix \ref{appendix:expdetail}.

\paragraph{Comparisons with Baselines.}
We compare our work with our baselines and other work on the multimodal sentence summarization dataset. Table \ref{tab:experiment} shows the results of different models. The results show that \textbf{UniG} performs comparably with  \textbf{UniG (T)}. \textbf{CFSum}s build on \textbf{UniG}, and introduces coarse-to-fine contribution network. ``F'', ``W'', and ``P'' represent the pre-filter, the word-level complement, and the phrase-level complement modules contained in the CFSum. The footnote is the location of the corresponding module. For example, CFSum-F$_3$ contains a pre-filter module with $L_f=3$. 
Generally, our methods \textbf{CFSum}s outperform the baselines \textbf{UniG (T)} and \textbf{UniG}. The best methods is \textbf{CFSum-F$_3$W$_6$P$_9$}. And it achieves 1.64 higher points on ROUGE-1 than \textbf{UniG}. 
We also conduct ablation experiments by applying one or two kinds of contributions. The results demonstrate that each image contribution benefits the model. In addition, combining all image contributions brings greater gains than a single contribution. Therefore, it can be concluded that different contributions are complementary to the summary. 
Besides, we conduct ablation studies by placing the pre-filter module at the beginning ($L_f=3$) or the end of the hierarchical layers ($L_f=9$). In comparison, placing the pre-filter module at the beginning (\textbf{CFSum-F$_3$W$_6$P$_9$}) yields better performance. 

\begin{table}[t]
\resizebox{\linewidth}{!}{%
\begin{tabular}{l|ccc}
\hline
\multicolumn{1}{c|}{Model} & Informativeness & Fluency & Non-Redundancy \\ \hline
UniG (T) & 3.63 & 3.48 & 2.91 \\
UniG & 3.69 & 3.66 & 3.05 \\
CFSum & \textbf{3.91} & \textbf{3.90} & \textbf{3.31} \\ \hline
\end{tabular}%
}
\caption{Human evaluations. 1 stands for the worst, and 5 stands for the best for three metrics.}
\label{tab:human}
\end{table}

\subsection{Human Evaluation Results}
We randomly select 50 samples from the test dataset and invite three postgraduates to score 1-5 for the summary quality. The evaluation metrics include informativeness, fluency, and non-redundancy. (1) Informativeness: Does the system summary contain comprehensive reference content? (2) Fluency: Is the system summary grammatically correct and readable? (3) Non-Redundancy: Does the system summary not have redundant or incorrect information relative to the reference summary? Table \ref{tab:human} shows the human evaluation results. We run the inter-annotator agreement study on three volunteers’ scores and achieve reasonable scores, 0.47, 0.39, and 0.43 on informativeness, fluency, and non-redundancy, respectively. The results show that our method \textbf{CFSum-F$_3$W$_6$S$_9$} achieves the best performance on all three aspects over \textbf{UniG (T)} and \textbf{UniG} baselines. Thus we conclude that our method is also effective through human evaluation.

\subsection{Further Analysis}
\label{para:analysis}
\subsubsection{Complement Modules Analysis}
In other multimodal tasks such as image captioning and multimodal translation, their models learn to attend to the image more for visual words like ``red'', ``rose'' and ``woman'' \citep{Lu2017KnowingWT, calixto-etal-2017-doubly}. Since our proposed complement modules aim at extracting complementary information relative to textual modality, we want to know which word or phrase the image provides gains on. As shown in Figure \ref{fig:analysis}, we visualize the complement gain value for the input words. We manually align the reference summary and the input text. The word highlighted with a red box indicates that it appears in the reference summary generatively\footnote{``Generatively'' means that the summary word is obtained by paraphrasing or synonymous substitution of the input word.} or extractively. 
\par First, we find that words with positive image gain can basically cover the reference summary information. It proves that our calculated gain helps in generating the target summary words. Second, it can be observed that different complement modules bring positive gains in different areas, which means different levels of complement modules are complementary. It further explains that multiple contributions are better than a single contribution in the experimental results. At last, it is worth noting that some words are gained from the image but are not visible in the image, \textit{i.e.}, ``relatives'' and ``victims''. Therefore, we believe the image brings gain in both visible and invisible words. We explain further in \nameref{sec:gainable}.

\begin{figure}[t]
    \centering
    \includegraphics[width=0.5\textwidth,scale=0.50]{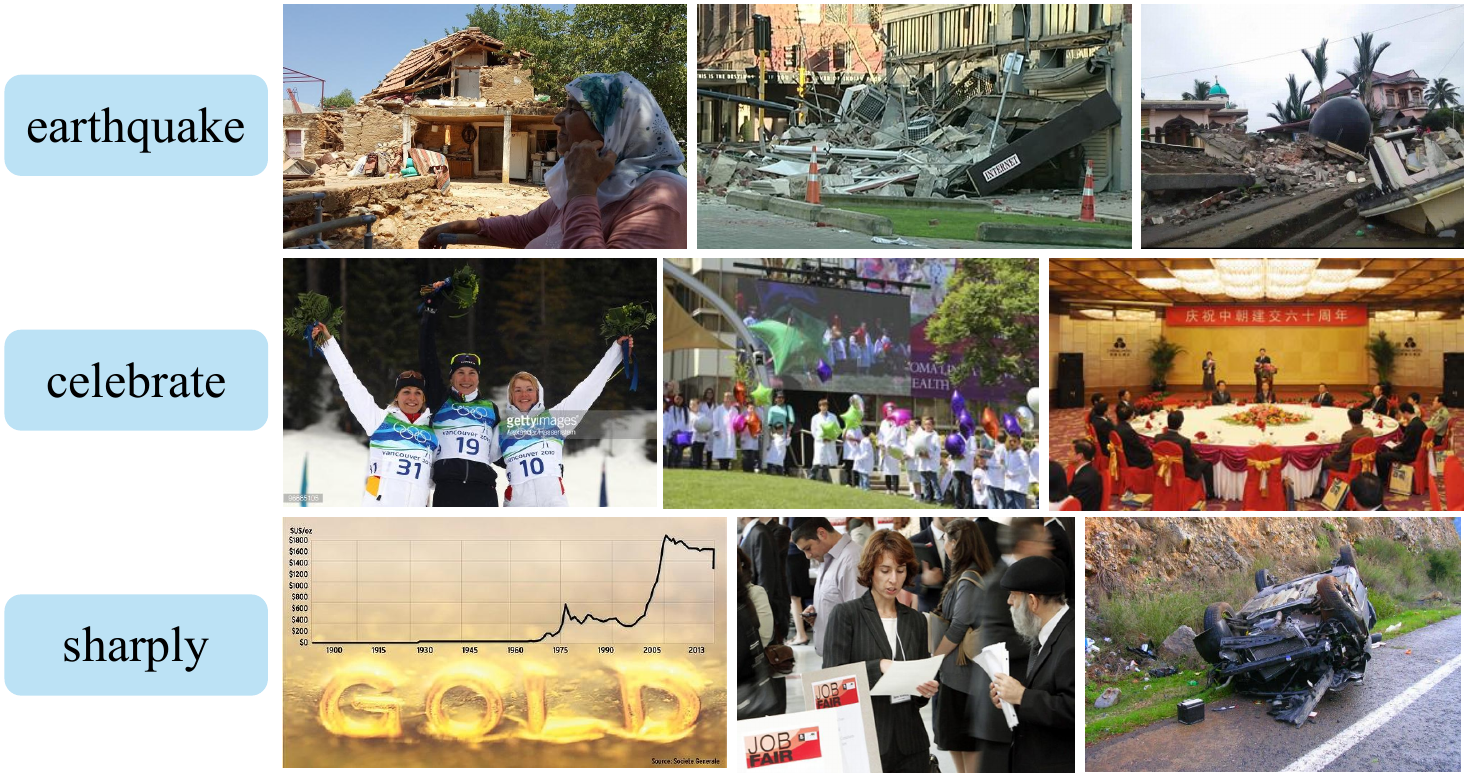}
    \caption{Visualization of gainable images.}
    \label{fig:visible}
\end{figure}

\subsubsection{Pre-filter Module Analysis}
Since we believe that images should provide meaningful contributions instead of robustness enhancements in multimodal summarization, we wonder whether unpaired multimodal data may affect the performance of our model. Therefore, we try generating the summary based on the unpaired image and text. 
\par In the test set, most of the images are highly similar in theme and content. Generating unpaired data with automatic shuffling is not significant for analysis. Therefore, we manually exchange $v_i$ and $v_j$ in pairs <$t_i,v_i$>, <$t_j,v_j$>, where $v_i,v_j$ have different themes or contents.
\par We exchange 20 pairs from 100 pairs of test samples. And we conduct experiments with different sampling for three times. The mean and standard deviation reports as Table \ref{tab:shuffle}. ``Paired'' represents ROUGE-1 on test set, ``Unpaired'' represents ROUGE-1 on the unpaired set. ``CFSum (filter-off)'' represents turning down the pre-filter mechanism. 
\begin{table}[h]
\centering
\scalebox{0.8}{%
\begin{tabular}{l|cc}
\hline
\multicolumn{1}{c|}{\textbf{Model}} & \textbf{Paired} & \textbf{Unpaired} \\ \hline
UniG & {\color[HTML]{000000} 46.22} & {\color[HTML]{000000} 46.20(±0.012)} \\
CFSum & {\color[HTML]{000000} 47.86} & {\color[HTML]{000000} 47.46(±0.007)} \\
CFSum (filter-off) & {\color[HTML]{000000} 47.77} & {\color[HTML]{000000} 47.12(±0.011)} \\ \hline
\end{tabular}%
}
\caption{Performance of unpaired multimodal data for the baseline and our methods.}
\label{tab:shuffle}
\end{table}%

\par The results show different trends in the two models. For UniG, unpaired multi-modalities do not affect the performance. We guess UniG does not exploit meaningful image information while relying only on text to generate the summary. In contrast, CFSum hurt more severely from unpairing. The difference exists because CFSum depends on the image and text. Thus, the unpaired image would reduce the correct information that CFSum gets. However, CFSum still performs better than UniG, proving that it is fault-tolerant. Furthermore, CFSum (filter-off) significantly suffers from unpaired data, showing that pre-filter can eliminate useless images.

\subsubsection{Ablation Study}
\begin{figure}[t]
    \centering
    \includegraphics[width=\linewidth,scale=1.00]{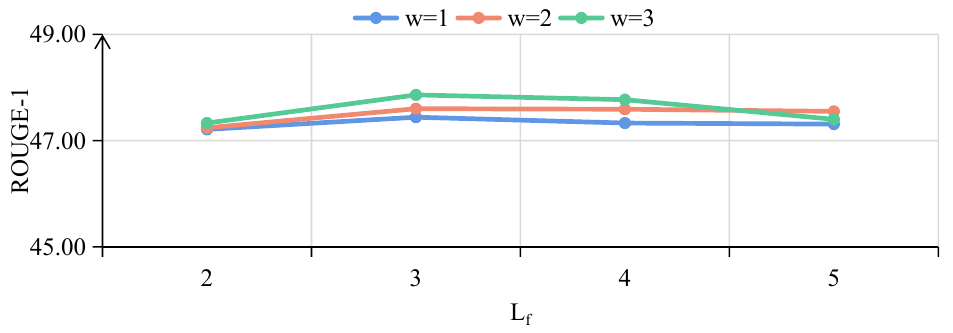}
    \caption{Ablation studies of layer setting.}
    \label{fig:layer}
\end{figure}

One of the most important hyperparameters in CFSum is the location of different contribution modules. Because the three modules' order in the network is fixed, we change their absolute position in the encoder layers and report the corresponding performance in Figure \ref{fig:layer}. $w$ denotes the number of layers between two modules, and the $X$ axis denotes the starting layer of the pre-filter module.
The results show that the different layer settings achieve comparable performance. It is noticeable that $w=2$ weakens the model. This is due to the fact that the network with small $w$ loses the advantage of a hierarchical structure in the encoder.

\subsubsection{Gainable Images}
\label{sec:gainable}
We select three gained words and corresponding gainable images to show in Figure \ref{fig:visible}. Consistent with our perception, images bring gains on visual words, such as ``earthquake''. More importantly, they bring gains on non-visual words such as ``celebrate'' and ``victims''. For example, ``celebrate'' may be used in competitions, events, and diplomacy as shown in Figure \ref{fig:visible}. Multimodal tasks such as image captioning or multimodal question answering focus on establishing associations between visual words and images. However, multimodal summarization also needs to pay attention to the associations between non-visual words and images. In other words, image contributes to both visual and non-visual words.

\section{Conclusion}
Based on the observation that existing multimodal summary models do not take full advantage of useful image information, this paper focuses on modeling different contributions of images for summarization. Therefore, we propose a novel framework CFSum consisting of pre-filter, word-level complement, and phrase-level complement modules. The pre-filter coarsely eliminates the impact of useless images. The two-level visual complement modules measure different aspects of image gains and guide the fusion of different modalities. Experimental results have shown that CFSum can significantly improve the summary. More importantly, the complement modules make images contribute to visual words and non-visual words.

\section*{Limitations}
Since our method constructs on the multimodal transformer, it cannot be migrated to the dual-stream model. Experiment results show that CFSum can achieve comparable performance with strong baselines. But it still cannot surpass the SOTA of some dual-stream large models.

\section*{Acknowledgements}
The research work has been supported by the Natural Science Foundation of China under Grant No. 62106263. 
\bibliography{custom}
\bibliographystyle{acl_natbib}

\newpage
\appendix
\section{Experiment details}
\label{appendix:expdetail}
Here, we will introduce some detailed settings for our experiments. All methods are run on NVIDIA GeForce RTX 3090. UniG has 139M parameters. When the batch size is 8, it takes 20 hours to train for 50 epochs with a single GPU.
\par We also provide evaluation scripts for reproduction. For ROUGE score, we use file2rouge\footnote{https://github.com/pltrdy/files2rouge} with default settings. For BERTScore\footnote{https://pypi.org/project/bert-score/0.2.1}, we use the official API. It exploits the pre-trained contextual embeddings from BERT to calculate the similarity between the hypothesis sentences and the reference sentences. For MoverScore, we use moverscore\_v2\footnote{https://github.com/AIPHES/emnlp19-moverscore}, which leverages BERT and Earth Mover Distance to measure the similarity.

\section{Derivation details}
\label{appendix:derivation}
Derivation detail of formula \ref{fomula:gi_log} is:
\begin{small}
    \begin{align}
        GI_j & = Gain(y_j/m^{L_w}, y_j/u^{L_w})\nonumber \\
            & = \textrm{KL}(\hat{y_j}||y_j/m^{L_w})-\textrm{KL}(\hat{y_j}||y_j/u^{L_w}) \nonumber \\
            & = P(\hat{y_j}=1)\cdot logP(y_j=1/m^{L_w}) \nonumber \\
            & + P(\hat{y_j}=0)\cdot logP(y_j=0/m^{L_w}) \nonumber \\
            & - P(\hat{y_j}=1)\cdot logP(y_j=1/u^{L_w}) \nonumber \\
            & - P(\hat{y_j}=0)\cdot logP(y_j=0/u^{L_w}) \nonumber \\
            & = P(y_j=\hat{y_j})\cdot logP(y_j=\hat{y_j}/m^{L_w}) \nonumber \\
            & - P(y_j=\hat{y_j})\cdot logP(y_j=\hat{y_j}/u^{L_w}) \nonumber \\
            & = logP(y_j=\hat{y_j}/m^{L_w})-logP(y_j=\hat{y_j}/u^{L_w}) 
    \end{align}
\end{small}%
Thus the gain is simplified to entropy difference.

\section{Examples of Complement Modules}
\label{appendix:exs}
We will provide some examples to explain further \nameref{sec:word-level}. For one of the input words $t_j$, we assume that it appears in the reference summary. Then the ground truth of the copy classification is $\hat{y_j}=1$. We list hypothetical classification results of bi-modal and uni-modal in Table \ref{tab:copy-results}.

\begin{table}[h]
\centering
\resizebox{0.5\columnwidth}{!}{%
\begin{tabular}{l|cc}
\hline
 & P($y_j=1$) & P($y_j=0$) \\ \hline
$u^{L_w}$ & 0.4 & 0.6 \\
$m^{L_w}$ & 0.6 & 0.4 \\ \hline
\end{tabular}%
}
\caption{Copy classification task results.}
\label{tab:copy-results}
\end{table}%
Then, the $GI_j$ is calculated as: 

\begin{small}
 \begin{align}
        GI_j & = Gain(y_j/m^{L_w}, y_j/u^{L_w})\nonumber \\
            & = logP(y_j=1/m^{L_w})-logP(y_j=1/u^{L_w}) \nonumber \\
            & = log0.6-log0.4 \nonumber \\
            & = 0.405
    \end{align}
\end{small}%
which means the image may give the input word $t_j$ a gain of 0.405. Furthermore, the image brings a positive gain. Thus in the attention layer, the text word $t_j$ should give the image a higher attention score.

\section{Impact of image category}
\label{appendix:dataset}
\begin{figure*}[t]
    \centering
    \includegraphics[width=\textwidth,scale=1.00]{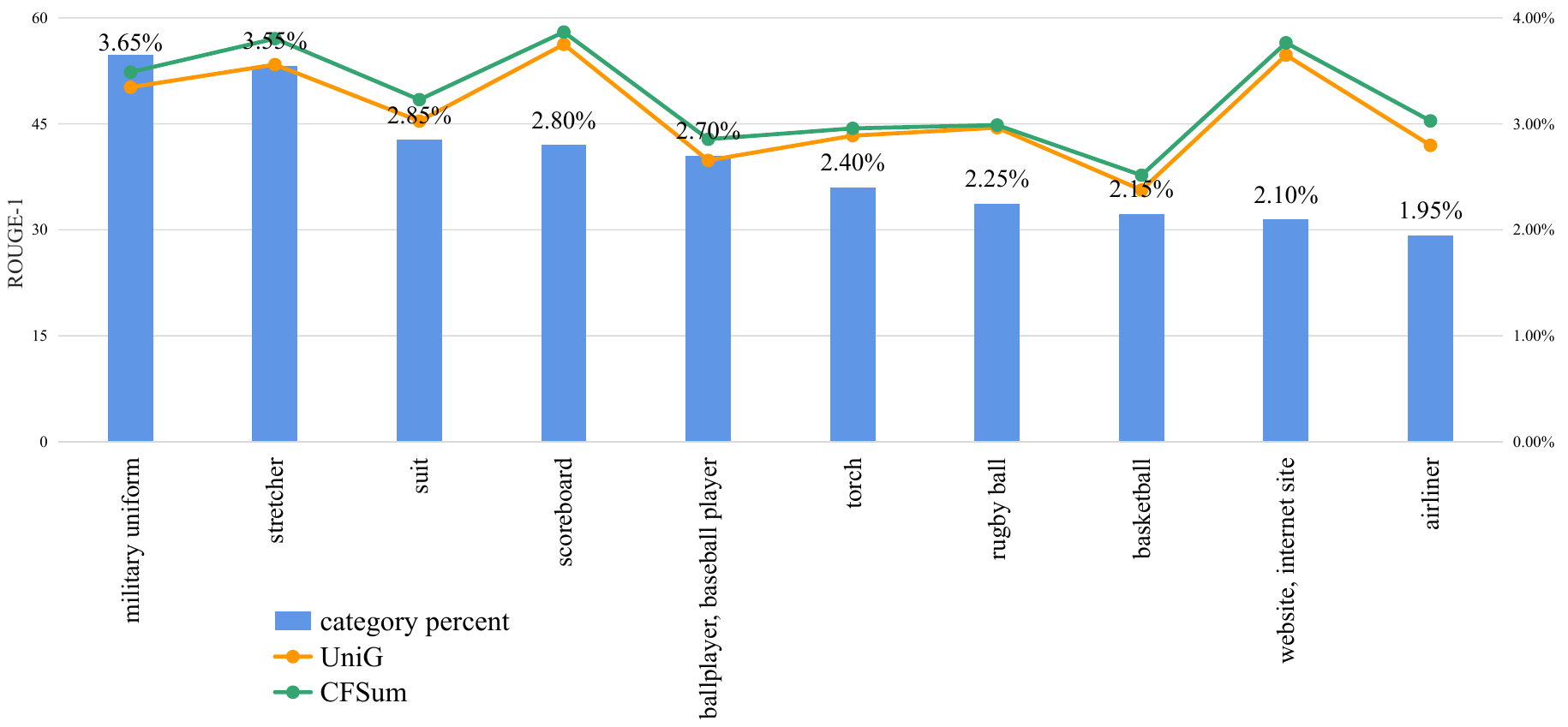}
    \caption{Top10 categories of test images and their corresponding performance.}
    \label{fig:category}
\end{figure*}

To further analyze the impact of our approach on different categories of images. We categorize the test images with VGG19 and show the performance of each type of image. As shown in Figure \ref{fig:category}, there are 380 categories in the test images, and we list the top 10 categories with the highest proportion. It can be seen that the image is evenly distributed. The line charts also show that CFSum is superior to UniG in all categories. Therefore there is no category bias in our method.

\section{Guided Attention}
\label{appendix:guided}
\begin{figure*}[h]
    \centering
    \setlength{\abovecaptionskip}{0.cm}
    \includegraphics[width=\textwidth,scale=1.00]{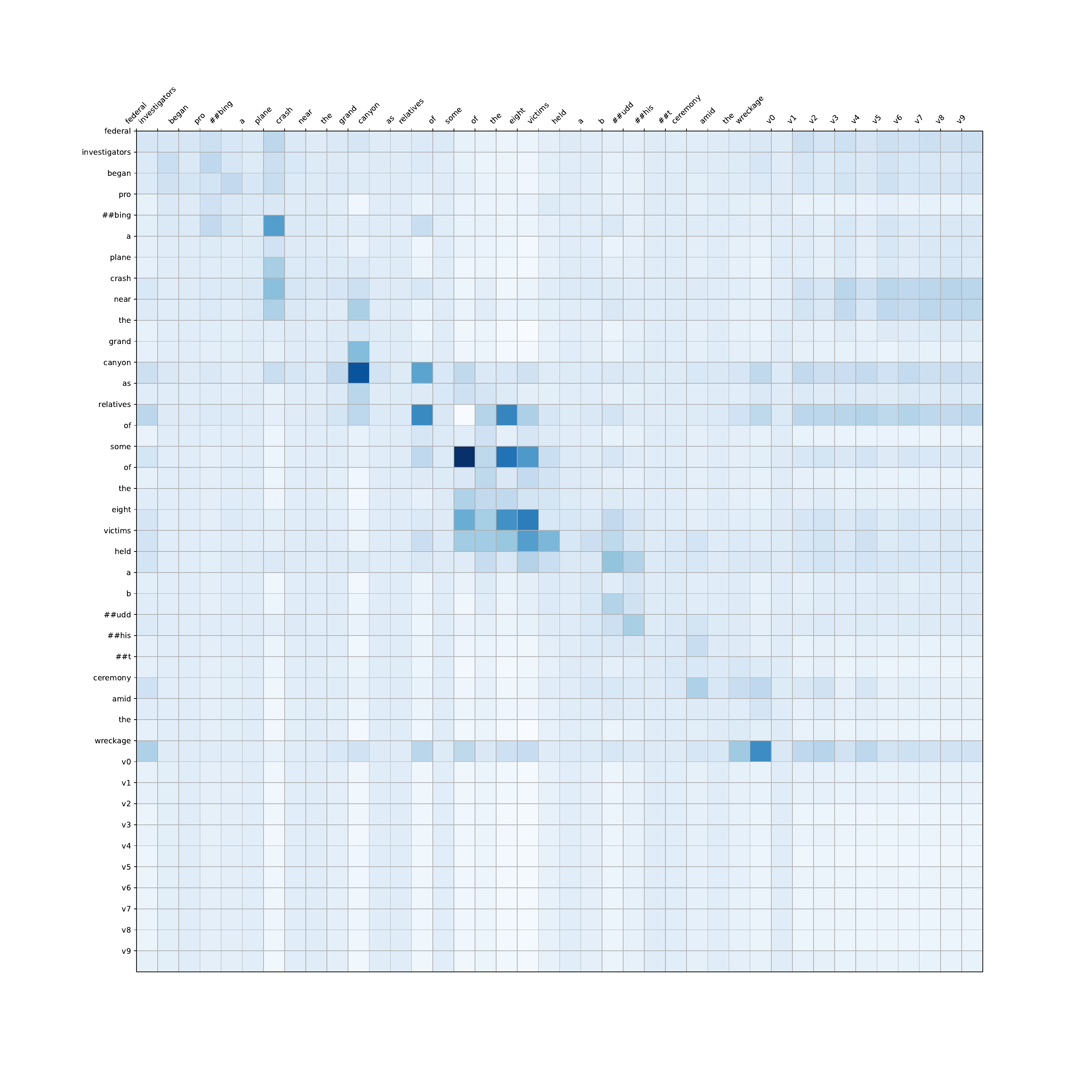}
    \caption{Visualize $8^{th}$ layer's attention under the word-level guided module. The reference summary is ``crash investigation begins relatives mourn eight victims''. $v_0\sim v_9$ is the image object detected feature.}
    \label{fig:wordAttn}
\end{figure*}

\begin{figure*}[h]
    \centering
    \setlength{\abovecaptionskip}{0.cm}
    \includegraphics[width=\textwidth,scale=1.00]{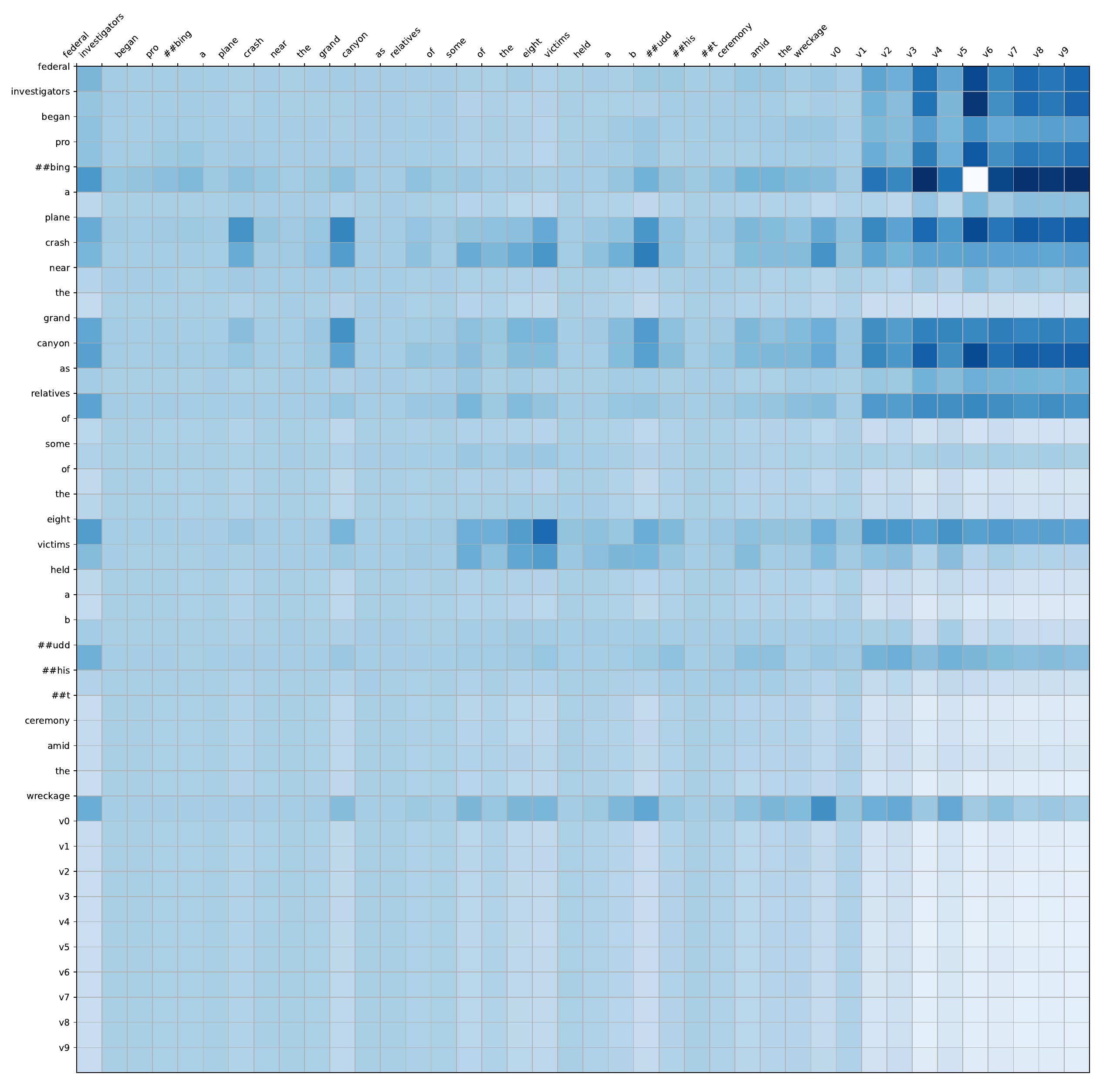}
    \caption{Visualize $11^{th}$ layer's attention under the word-level guided module. The reference summary is ``crash investigation begins relatives mourn eight victims''. $v_0\sim v_9$ is the image object detected feature.}
    \label{fig:phraseAttn}
\end{figure*}

We visualize (1) the attention matrix from the $8^{th}$ encoder layer of CFSum-F$_3$W$_6$S$_9$, whose layer is under the word-level guidance. (2) the attention matrix from the $11^{th}$ encoder layer of CFSum-F$_3$W$_6$S$_9$, whose layer is under the phrase-level guidance. The attention matrix is renormalized after removing [CLS] and [SEP]. They are shown in Figure \ref{fig:wordAttn} and Figure \ref{fig:phraseAttn}.
\par From the attention under the word-level guidance, we can observe that some input words which generatively or extractively occur in the reference summary will attend to the image, such as ``crash'' and ``relatives''. From the attention under the phrase-level guidance, we can observe that some input phrases which generatively or extractively occur in the reference summary attend to the image more. Above all, it also proves that two visual complement modules succeed in providing better encoding to generate summaries.

\end{document}